\documentclass{article}

\usepackage{microtype}
\usepackage{graphicx}
\usepackage{subfigure}
\usepackage{booktabs} 

\usepackage{hyperref}

\usepackage[accepted]{icml2018_no_icml}

\icmltitlerunning{A Multi-Horizon Quantile Recurrent Forecaster}

\def\tightlist{}
\usepackage{graphicx}
\usepackage{amssymb,amsmath}
\usepackage{longtable}

\begin{document}

\twocolumn[
\icmltitle{A Multi-Horizon Quantile Recurrent Forecaster}

\begin{icmlauthorlist}
\icmlauthor{Ruofeng Wen}{fcst}
\icmlauthor{Kari Torkkola}{fcst}
\icmlauthor{Balakrishnan Narayanaswamy}{ai}
\icmlauthor{Dhruv Madeka}{fcst}
\end{icmlauthorlist}

\icmlaffiliation{fcst}{Forecasting Data Science, Amazon}
\icmlaffiliation{ai}{Amazon AI Lab, Amazon}

\icmlcorrespondingauthor{Ruofeng Wen}{ruofeng@amazon.com}
\vskip 0.3in
]

\printAffiliationsAndNotice{} 

\begin{abstract}
We propose a framework for general probabilistic multi-step time series
regression. Specifically, we exploit the expressiveness and temporal
nature of Sequence-to-Sequence Neural Networks (e.g.~recurrent and
convolutional structures), the nonparametric nature of Quantile
Regression and the efficiency of Direct Multi-Horizon Forecasting. A new
training scheme, \emph{forking-sequences}, is designed for sequential
nets to boost stability and performance. We show that the approach
accommodates both temporal and static covariates, learning across
multiple related series, shifting seasonality, future planned event
spikes and cold-starts in real life large-scale forecasting. The
performance of the framework is demonstrated in an application to
predict the future demand of items sold on Amazon.com, and in a public
probabilistic forecasting competition to predict electricity price and
load.
\end{abstract}

\section{Introduction}\label{introduction}

Classical time series forecasting models aim to predict \(y_{t+1}\)
given recent history \(y_{:t} = (y_t,\cdots,y_0)\). Common approaches
include Box-Jenkins method, i.e.~ARIMA models
(\href{https://books.google.com/books?hl=en\&lr=\&id=rNt5CgAAQBAJ\&oi=fnd\&pg=PR7\&ots=DJ18wTl-UA\&sig=yyMLboZO6Q7uTQRxF5rMar1gnsE\#v=onepage\&q\&f=false}{Box
et al, 2015}). In practice, forecasting problems are far more complex.
Many related time series are present. Inputs involve multiple covariates
such as dynamic historical features, static attributes for each series
and known future events. Series often have long term dependency such as
yearly seasonality pattern, with nonlinear relationships between inputs
and outputs. Usually, multi-step, long-horizon forecasts are needed,
together with precise prediction intervals to quantify forecast
uncertainties required to estimate risks in decision making. Modern
methods have been proposed to attack these issues \emph{individually}.

Recurrent Neural Networks (RNN,
\href{http://psych.colorado.edu/~kimlab/Elman1990.pdf}{Elman, 1990})
have recently demonstrated state-of-art performance in various
applications. An RNN learns a fixed-length nonlinear representation from
multiple sequences of arbitrary length. Historically, RNN fits into the
Nonlinear Autoregressive Moving Average framework
(\href{https://pdfs.semanticscholar.org/0677/8bd87125a28f0d045e0221ca1b8ad1d469b6.pdf}{Connor
et al, 1992}). The most popular variant, Long-Short-Term-Memory networks
(LSTM,
\href{http://digital-library.theiet.org/content/conferences/10.1049/cp_19991218}{Gers
et al, 1999}) were designed to cope with the vanishing gradient problem,
which is essential to capturing long-term dependency.
\href{https://arxiv.org/pdf/1308.0850.pdf}{Graves, 2013} introduced
Sequence-to-Sequence RNN (Seq2Seq) with the ability to generate a future
sequence, usually a sentence, given the previous one. Such architecture
is intimately related to multi-step time series forecasting, a
connection which has been well investigated in recent studies
(\href{https://arxiv.org/pdf/1703.10089.pdf}{Cinar et al, 2017} and
\href{https://arxiv.org/pdf/1704.04110.pdf}{Flunkert et al, 2017}).
Notably, Convolutional neural networks (CNN) with Seq2Seq structures
have gained recent popularity, after the success of WaveNet
(\href{https://arxiv.org/pdf/1609.03499.pdf}{Van Den Oord et al, 2016})
in the field of audio generation, and have also been studied under the
topic of forecasting
(\href{https://arxiv.org/pdf/1703.04691.pdf}{Borovykh et al, 2017}).

Most applications of Neural Networks to time series, including Seq2Seq
with both RNN and CNN, build on one approach: they train a model to
predict the one-step-ahead estimate \(\hat{y}_{t+1}\) given \(y_{:t}\),
and then iteratively feed this estimate back as the ground truth to
forecast longer horizons. This is knowns as the \emph{Recursive}
strategy to generate multi-step forecasts, also sometimes referred to as
\emph{iterative} or \emph{read-outs} in literature. Due to its similar
form to auto-regressive or Markovian assumptions in modeling, the
Recursive strategy is usually taken for granted.
\href{https://papers.nips.cc/paper/5956-scheduled-sampling-for-sequence-prediction-with-recurrent-neural-networks.pdf}{Bengio
et al, 2015} and \href{https://arxiv.org/pdf/1610.09038.pdf}{Lamb et al,
2016} pointed out that a carefully designed training scheme is needed
when the Recursive strategy is applied with RNN, to avoid the
discrepancy between consuming actual data versus estimates during
prediction, since the latter leads to error accumulation. In the field
of forecasting,
\href{http://www.vwl.tuwien.ac.at/hanappi/AgeSo/rp/Chevillon_2007.pdf}{Chevillon,
2007} showed that the Direct strategy, where a model directly predicts
\(y_{t+k}\) given \(y_{:t}\) for each \(k\), is less biased, more stable
and more robust to model mis-specification. A comprehensive comparison
by \href{http://alumnus.caltech.edu/~amir/biasvar-fin.pdf}{Taieb and
Atiya, 2016} investigated different multi-step strategies with Neural
Networks, and recommended the Direct Multi-Horizon strategy: directly
train a model with a multivariate target \((y_{t+1},\cdots,y_{t+k})\).
The Multi-Horizon strategy avoids error accumulation, yet retains
efficiency by sharing parameters .

Many decision making scenarios require the richer information provided
by a probabilistic forecast model that returns the full conditional
distribution \(p(y_{t+k}|y_{:t})\), rather than a point forecast model
that predicts only the conditional mean \(\mathbb{E}(y_{t+k}|y_{:t})\).
A canonical example is a task with asymmetric costs for over and
under-prediction. Then the symmetric Mean Squared Error, which the
conditional mean minimizes, does not reflect the true loss. For
real-valued time series, probabilistic forecast is traditionally
achieved by assuming an error distribution or stochastic process,
usually Gaussian, on the residual series
\(\epsilon_t = y_t - \hat{y_t}\). However, an exact parametric
distribution is often not directly relevant in applications. Instead,
particular quantiles of the forecast distribution are useful in making
optimal decisions, both to quantify risks and minimize losses (e.g.~risk
management, power grid capacity optimization), leading to the use of
Quantile Regression (QR,
\href{https://pdfs.semanticscholar.org/a3cd/bfbba2ef3ce285980edc1213a4ac56f05bb1.pdf}{Koenker
and Gilbert, 1978}). QR learns to predict the conditional quantiles
\(y_{t+k}^{(q)}|y_{:t}\) of the target distribution, i.e.
\(\mathbb{P}(y_{t+k}\leq y_{t+k}^{(q)}|y_{:t}) = q\). QR is robust since
it does not make distributional assumptions, produces accurate
probabilistic forecasts with sharp prediction intervals, and often
serves as a post-processor for prediction calibration
(\href{http://citeseerx.ist.psu.edu/viewdoc/download?doi=10.1.1.470.5844\&rep=rep1\&type=pdf}{Taylor,
2000}).

To reconcile and improve upon these separate methods, we propose
MQ-R(C)NN: a Seq2Seq framework that generates Multi-horizon Quantile
forecasts. The model is designed to solve the large scale time series
regression problem:

\[p(y_{t+k,i},\cdots,y_{t+1,i}|y_{:t,i},x_{:t,i}^{(h)},x_{t:,i}^{(f)},x_{i}^{(s)})\]

where \(y_{\cdot,i}\) is the \(i\)th time series to forecast,
\(x_{:t,i}^{(h)}\) are the temporal covariates available in history,
\(x_{t:,i}^{(f)}\) is the knowledge about the future, and
\(x_{i}^{(s)}\) are the static, time-invariant features. Each series is
considered as one sample fed into a single RNN or CNN, even if they
correspond to different items. This enables cross-series learning and
cold-start forecasting for items with limited history. For readability,
the sample/series subscript \(i\) will be dropped from now on.

To our best knowledge, this is the first work to combine sequential nets
like RNNs and one-dimensional CNNs with either QR or Multi-Horizon
forecasts. We demonstrate in details how the individual attributes of
each methods combine seamlessly in the framework, and achieve better
performance than state-of-art models in multiple forecasting
applications. The major contributions of this paper also include:

\begin{itemize}
\tightlist
\item
  We propose an efficient training scheme for the combination of
  sequential neural nets and Multi-Horizon forecast. The approach, which
  we call \emph{forking-sequences} and detailed in
  \protect\hyperlink{training-scheme}{Section 3.3}, can dramatically
  improve training stability and performance of encoder-decoder style
  recurrent nets or ConvNets, by training on all time points where a
  forecast would be created, in a one pass over the data series.
\item
  We design a network sub-structure to accommodate a previously
  little-attended issue: how to account for known future information,
  including the alignment of shifting seasonality and known events that
  cause large spikes and dips.
\end{itemize}

The rest of this paper is organized as follows. In
\protect\hyperlink{related-work}{Section 2} we discuss prior work, and
highlight the novel aspects of our work. In
\protect\hyperlink{method}{Section 3}, we describe our proposed
MQ-R(C)NN framework in detail, together with variants that we have found
useful in practice. In particular, we describe the generality and how
different sequential structures can be used fruitfully in practice. In
\protect\hyperlink{application}{Section 4}, we demonstrate the value of
MQ-R(C)NN on a large dataset of retail demand time series from Amazon,
and on data from a public electricity forecasting competition, where we
beat the state of the art. \protect\hyperlink{conclusion}{Section 5}
draws some conclusions and outlines possible directions for future
research.

\hypertarget{related-work}{\section{Related Work}\label{related-work}}

RNNs and CNNs have been recently applied to time series \emph{point}
forecasting.
\href{http://www.diva-portal.org/smash/get/diva2:710518/FULLTEXT02}{Längkvist
et al, 2014} reviewed on time series modeling with deep learning in
various fields of study.
\href{https://arxiv.org/pdf/1705.04378.pdf}{Bianchi et al, 2017}
presented a comparative study on the performance of various RNNs applied
to the Short Term Load Forecasting problem.
\href{https://arxiv.org/pdf/1703.10089.pdf}{Cinar et al, 2017}
investigated the attention model for Seq2Seq on both univariate and
multivariate time series.
\href{https://arxiv.org/pdf/1703.04691.pdf}{Borovykh et al, 2017}
applied dilated CNNs on financial time series. However, these efforts
are all built on the Recursive strategy.
\href{http://alumnus.caltech.edu/~amir/biasvar-fin.pdf}{Taieb and Atiya,
2016} analyzed the performance of different multi-step strategies on a
Multi-Layer Perceptron (MLP), where the Direct Multi-Horizon strategy
stands out.

For \emph{probabilistic} forecasting with encoder-decoder models,
\href{https://arxiv.org/pdf/1704.04110.pdf}{Flunkert et al, 2017}
propose DeepAR, a Seq2Seq architecture with an identical encoder and
decoder. DeepAR directly outputs parameters of a Negative Binomial. This
is similar to \href{https://arxiv.org/pdf/1702.05386.pdf}{Ng et al,
2017} where an MLP predicts Gaussian parameters, and such a strategy
dates back to
\href{https://www.microsoft.com/en-us/research/wp-content/uploads/2016/02/bishop-ncrg-94-004.pdf}{Bishop,
1994}. DeepAR is trained by maximizing likelihood and Teacher Forcing
(feeding ground truth recursively in training), and during prediction
time it is fed a sample drawn from the estimated parametric
distribution. This sampling is performed multiple times to generate a
series of \emph{sample paths}, as the empirical distribution of
forecasts. Our method differs from DeepAR by using the more practically
relevant Multi-Horizon strategy, a more efficient training strategy and
directly generating accurate quantiles.

For quantile forecasts with neural nets,
\href{http://citeseerx.ist.psu.edu/viewdoc/download?doi=10.1.1.470.5844\&rep=rep1\&type=pdf}{Taylor,
2000} used an MLP to generate quantile forecasts for financial returns.
The model was used to process the innovations of another GARCH model, to
obtain calibrated Value-at-Risk.
\href{http://bura.brunel.ac.uk/bitstream/2438/13197/1/Fulltext.pdf}{Xu
et al, 2016} designed a quantile autoregressive neural net for stock
price prediction. Instead of feeding the mean estimate or a sampled
instance, they fed previously estimated quantiles into the model.
Neither of the approaches used sequantial nets and exploit their
temporal nature. The former depends on an external model while in the
latter feeding back in quantiles is difficult to justify.

\hypertarget{method}{\section{Method}\label{method}}

In this section, we describe the loss function, neural network
architecture, how the network is trained, encoder extensions to further
enhance model performance and some practical consideration in the design
of of input features.

\subsection{Loss Function}\label{loss-function}

In Quantile Regression, models are trained to minimize the total
Quantile Loss (QL):

\[L_q(y,\hat{y}) = q(y-\hat{y})_{+} + (1-q)(\hat{y}-y)_{+}\]

where \((\cdot)_{+}=\max(0,\cdot)\). When \(q=0.5\), the QL is simply
the Mean Absolute Error, and its minimizer is the median of the
predictive distribution. Let \(K\) be the number of horizons of
forecast, \(Q\) be the number of quantiles of interest, then the
\(K\times Q\) matrix \(\hat{\mathbf{Y}} = [\hat{y}_{t+k}^{(q)}]_{k,q}\)
is the output of a parametric model \(g(y_{:t},x,\theta)\), e.g.~an RNN.
The model parameters are trained to minimize the total loss,
\(\sum_t \sum_q \sum_k L_q(y_{t+k},\hat{y}_{t+k}^{(q)})\), where \(t\)
iterates through all forecast creation times (FCTs). Depending on the
problem, components of the sum can be assigned different weights, to
highlight or discount different quantiles and horizons.

\subsection{Network Architecture}\label{network-architecture}

\begin{figure}
\centering
\includegraphics[width=0.50000\textwidth]{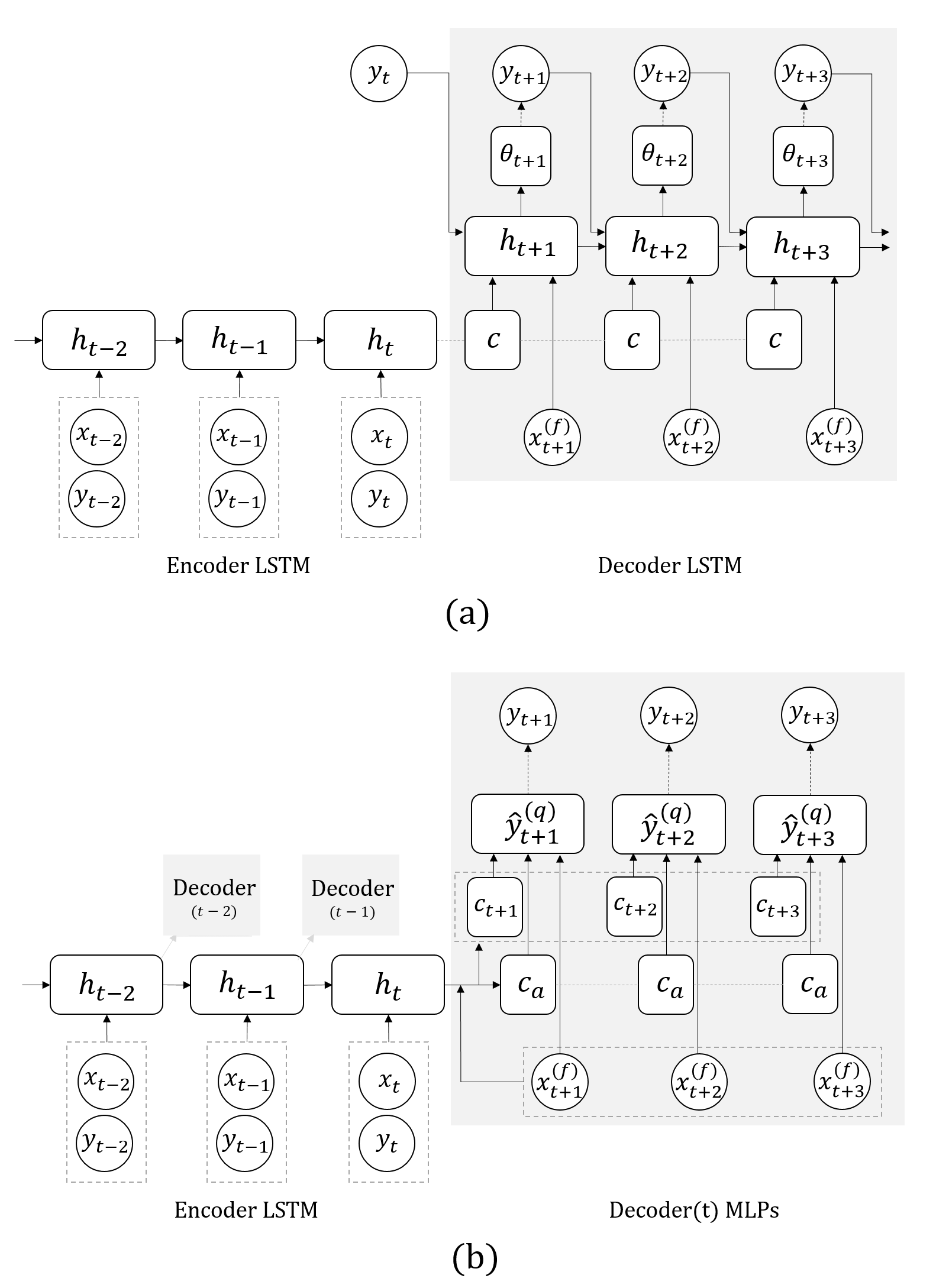}
\caption{Neural net architectures for multi-step forecasts. Circles and
squares denote observed and hidden nodes, respectively. Dashed box
flattens nodes into a vector. Dashed line means replication. Dashed
arrow is the loss, which links network output and targets.
\(x_t = (x_{t}^{(h)},x_{t}^{(f)},x^{(s)})\). Layer depth is not shown.
\textbf{(a)} Seq2SeqC, where the loss function is likelihood
(e.g.~Multinomial for text generation, Gaussian for numeric values),
parameterized by \(\theta_t\). At prediction time, \(\hat{y}_{t+k}\) is
fed into decoder, instead of \(y_{t+k}\) as in training. \textbf{(b)}
MQ-RNN, where the total loss function is sum of individual quantile
loss, and the output is all the quantile forecasts for different values
of \(q\). During training, the time sequence is forked: there is a
decoder corresponding to each recurrent layer with identical weights
(shaded boxes).}
\end{figure}

For simplicity, we consider the design of an RNN Seq2Seq model in this
section. The MQ-RNN architecture resembles the Seq2Seq with context
(Seq2SeqC, Figure 1a) proposed by
\href{https://arxiv.org/pdf/1406.1078.pdf}{Cho et al, 2014}. We here
also use a vanilla LSTM to encode all history into hidden states
\(h_t\). Instead of using an LSTM as the recursive decoder, MQ-RNN has a
design of two MLP branches. The first (\emph{global}) MLP summarize the
encoder output plus all future inputs into two contexts: a series of
horizon-specific contexts \(c_{t+k}\) for each of the \(K\) future
points, and a horizon-agnostic context \(c_a\) which captures common
information:

\[ (c_{t+1},\cdots,c_{t+K},c_a) = m_G(h_t,x_{t:}^{(f)}) \]

where \(m_G(\cdot)\) is the global MLP and contexts \(c_{(\cdot)}\) each
can have arbitrary dimension. The second (\emph{local}) MLP applies to
each specific horizon. It combines the corresponding future input and
the two contexts from the global MLP described earlier, then outputs all
the required quantiles for that specific future time step:

\[ (\hat{y}_{t+k}^{(q_1)},\cdots,\hat{y}_{t+k}^{(q_Q)}) = m_L(c_{t+k},c_a,x_{t+k}^{(f)}) \]

where \(m_L(\cdot)\) is the local MLP with its parameters shared across
all horizons \(k\in \{1,\cdots,K\}\), and \(q_{(\cdot)}\) denotes each
of the \(Q\) quantiles. The overall structure is illustrated in Figure
1b.

The local MLP is the key to aligning future seasonality and events and
the capability to generate sharp spiky forecasts. Since the parameters
are shared across horizons, it is tempting to replace it with another
(bidirectional) LSTM. However, this is unnecessary and expensive: the
flow of latent temporal information has already been captured by the
Direct Multi-Horizon-specific context. Furthermore, feeding predictions
recursively as surrogate of ground truth is not recommended since the
corresponding quantile outputs are non-additive and combining them
requires learning complicated functions.

At first glance, the two types of global context seem redundant. We
argue that the horizon-specific context is always necessary: it carries
\emph{network-structural} awareness of the temporal distance between a
forecast creation time point and a specific horizon. This is essential
to aspects like seasonality mapping. In Seq2SeqC, only horizon-agnostic
context exists, and horizon awareness is indirectly enforced by
recursively feeding predictions into the cell for the next time step.
The horizon-agnostic context is still included in our model, based on
the idea that not all relevant information is time-sensitive.
Empirically, we find that adding this structure to the model improves
the stability of learning and the smoothness of generated forecasts. In
cases where there is no meaningful future information, or sharp and
spiky forecasts is not desired, the local MLP can be removed, and a
simplified global MLP with
\(\text{vec}(\hat{\mathbf{Y}}) = m_G(h_t,x_{t:}^{(f)})\) still retains
all other advantages described above.

\hypertarget{training-scheme}{\subsection{Training
Scheme}\label{training-scheme}}

One major performance gain of our model over Seq2Seq is achieved by the
\emph{forking-sequences} training scheme we describe below. Note that
all Seq2Seq style models put an \emph{end} to the input sequence, e.g.~a
stopping symbol in natural language, and that end point is where encoder
and decoder exchange information. In forecasting, this stopping symbol
is naturally a \emph{forecast creation time} (FCT), the time step at
which a forecast for future horizons must be generated. Unlike many
other sequential modeling problem, time series forecasts often need to
be generated at each possible time point, i.e.~a forecast is required
each day or week. Most applications use \emph{cutting-sequences}: split
the time series at a set of randomly chosen FCTs and use each series/FCT
pair as a training example. This is also known as \emph{moving-window}
scheme in forecasting, and requires substantial data augmentation.

Such method is not necessary in an RNN thanks to its temporal nature. As
illustrated in Figure 1b, our framework creates Multi-Horizon forecasts
by placing a series of decoders, with shared parameters, at \emph{each}
recurrent layer (time point) in the encoder, and computes the loss
against the corresponding targets (future series relative to that time
point; can be populated on-the-fly in implementation). Thus we planted
the nature of forecasting-at-each-time application \emph{structurally}
into the neural net training. Then one back-propagation-through-time can
gather the multi-horizon error gradients of different FCTs in one pass
over a sample, with little additional cost. In the MQ-RNN example,
forking can be expressed mathematically as: \(\forall t\),
\(h_t = \text{encoder}(x_{:t},y_{:t})\),
\(\hat{y}^{(q)}_{t:} = \text{decoder}(h_t,x^{(f)}_{t:})\), where
\(\text{encoder}(\cdot)\) is an LSTM and \(\text{decoder}(\cdot)\) is
the global/local MLPs discussed in the previous subsection, and the
parameters of both are invariant of \(t\).

As a result of forking-sequences, each time series of arbitrary length
serves as a single sample in our model training, eliminating the need of
data augmentation, and dramatically reducing the training time. Note
that the prediction tasks at each FCT are highly correlated, so by
updating the gradients together, the optimization process is stabilized.
Empirically, this training scheme greatly boosts model performance and
regularizes learning stability by efficiently using all information in
one shot, while previous algorithms need to cut and down-sample data.
The benefit behind forking-sequences may also be related to ideas
described in \href{https://arxiv.org/pdf/1511.03677.pdf}{Lipton et al,
2015}, where a scalar categorical target is replicated to each recurrent
layer in a time series classification problem. Our approach differs by
utilizing the nature of the multi-step time series prediction problem to
implement the \emph{actual} forecasting task at each time point, and
thus enable the recurrent layers to convey both concepts of observed
time points and forecast creation time.

The Direct strategy is often criticized as not being able to use the
data between \(T-K\) and \(T\), where \(T\) is the end of training
period, since the Multi-Horizon target is not available beyond \(T\). We
resolve this issue by masking all the error terms after that point, so
the model can still learn shared parameters from the available
short-horizon partial targets when near the boundary of training period.
This target masking strategy is a general approach to remove any cases
when a (part of) multi-horizon forecast is unwanted or shouldn't be
evaluated, depending on application specifics.

\hypertarget{encoder-extensions}{\subsection{Encoder
Extensions}\label{encoder-extensions}}

In previous sections, the core design of MQ-RNN was described: a series
of multi-horizon, future-aligned forked decoders that forecast
quantiles. Here we discuss some practical extensions to the encoder, to
go beyond a vanilla LSTM and further improve performance. An
illustration of the structures described below can be found in Figure 2

\begin{figure}
\centering
\includegraphics[width=0.50000\textwidth]{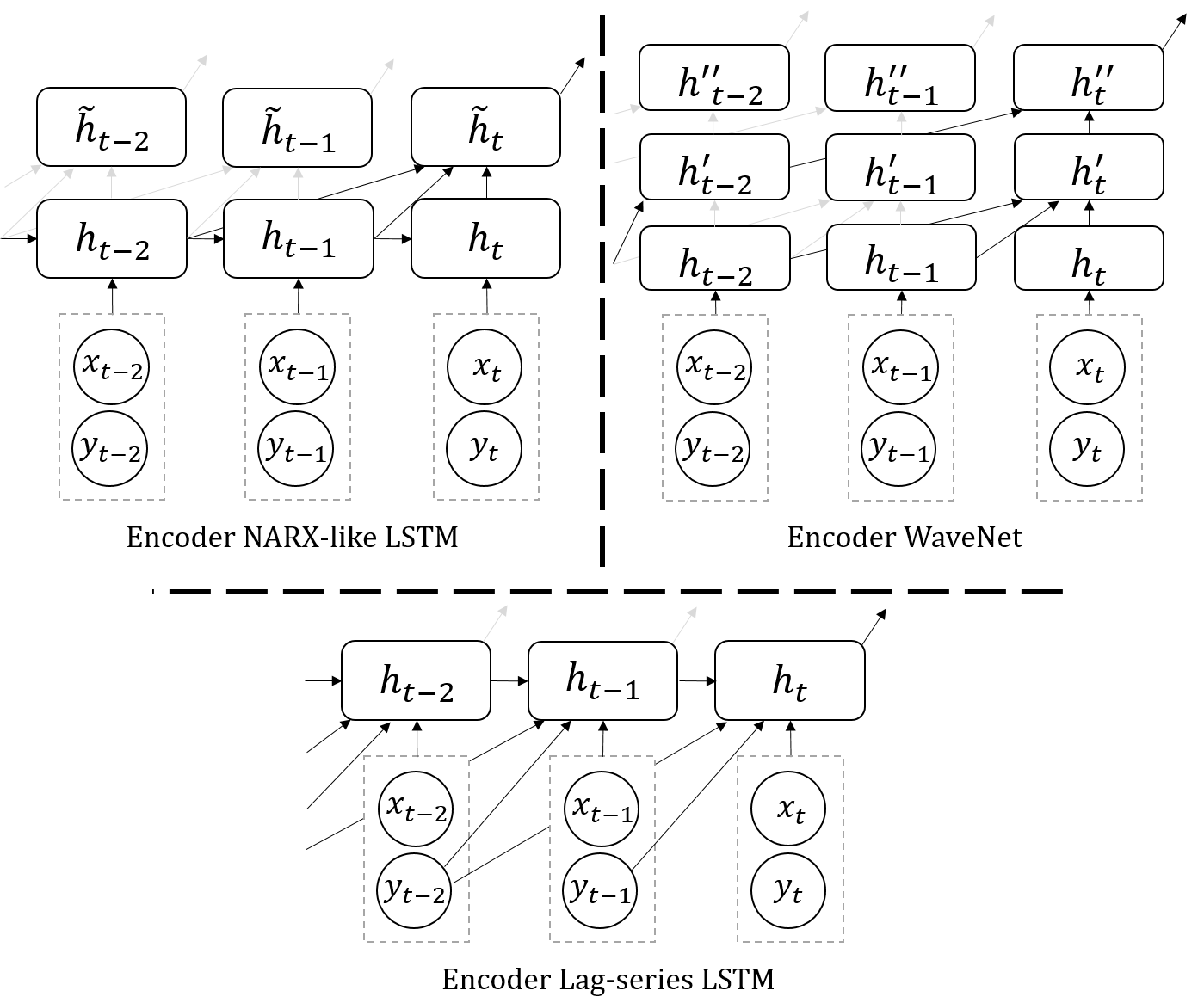}
\caption{Alternative Encoders for MQ-RNN (compare to and contrast with
Figure 1b). For clarity, all forking decoders are not shown, and
connections that do not contribute to the last decoder are in gray. Note
that \(h_t\) in LSTM encoders are from LSTM cells, while in WaveNet all
hidden states are from dilated causal convolutions.}
\end{figure}

LSTMs were proposed to avoid gradient vanishing and expand the long-term
memory capacity of RNNs. Many forecasting problems have long periodicity
(e.g.~365 days) and may suffer from memory loss during recurrent forward
propagation. Another not as well-known type of recurrent net to solve
the same long dependency issue is NARX RNN
(\href{https://arxiv.org/pdf/1702.07805.pdf}{DiPietro et al, 2017}),
which computes hidden state \(h_t\) not only based on \(h_{t-1}\), but
also a specific set of other past states, e.g.
\((h_{t-2},\cdots,h_{t-D})\). This is also known as
\emph{skip-connections}. The presence of past states reduces the
requirement on RNN cell's ability to memorize long dependencies. A
simple modification in MQ-RNN to enable a NARX-ish encoder is to put an
extra linear layer on top of the LSTM to summarize past states:
\(\tilde{h}_t = m(h_{t},\cdots,h_{t-D})\), and then feed \(\tilde{h}_t\)
instead into the global MLP decoder. Note this operation is compatible
with the forking-sequences optimization by placing the same \(m(\cdot)\)
on each recurrent layer and its trailing states.

The NARX-ish encoder does bring improvement over vanilla LSTM in
experiments. But a seemly naive alternative performs even better: just
feed past series \((y_{t-1},\cdots,y_{t-D})\) as lagged feature inputs,
along with \(y_t\), into the recurrent layer at \(t\). In fact, this
effectively constructs skip-connections to past values of the input
series before passing them through the recurrent layer, as opposed to
having skip-connections after RNN. The reason why this simple lag-series
trick works well could be due to the nature of forecasting: the
historical values of the time series is the most predictive information
we have for it's future values, and thus have the most influence on the
hidden states. Therefore the lag-series can better approximate a real
NARX encoder with \(h_t\) being updated by the distant past.

The choice of encoder is not restricted to recurrent networks. Any
neural net that has sequential or temporal structure and is compatible
with forking-sequences, can serve as an encoder in the MQ-framework.
\href{https://arxiv.org/pdf/1609.03499.pdf}{Van Den Oord et al, 2016}
proposed WaveNet to process and generate audio sequences, with a stack
of dilated causal 1D convolution layers. The higher-level dilated
convolution layers can reach far into the past summary in lower levels,
acting as another alternative of direct long-term memory connection.
Since these convolution kernels have stride 1 and the local structure is
step-invariant to allow forking-sequences, a WaveNet or stacked dilated
convolutional encoder can be seamlessly plugged into our model framework
(i.e.~MQ-CNN) and yield excellent forecasting performance in our
experiments, as shown in \protect\hyperlink{application}{Section 4.1}.

\subsection{Future and Static
Features}\label{future-and-static-features}

There are typically two kinds of known future information. Seasonal
features are simply (linear) kernels centered at a specific day of the
week, a moving holiday or any other seasonality labels. They are
commonly used in Generalized Additive Models for time series. Event
features are binary or numeric temporal indicators of if and how a
certain type of event happens (e.g.~price adjustment, censoring). If
these events are sufficiently frequent in training data, the model can
learn their effects from data and generate sharp changes in forecasts.
If the event can be planned (e.g.~promotion campaign), the model can
simulate its effect for decision making. In practice, we found that
distant future information (e.g.~a holiday) can have retrospective
impact on near future horizons (anticipation), which is why the global
MLP also uses future summaries.

Static features contain series-specific information. For instance, it
could be the sector of a stock, image and text description of a product,
or location of a power plant. In our experience, static features are
usually less predictive than time series ones, but combined with
training one model on multiple series, they bridge different sets of
time series behaviors and allow the model to borrow statistical strength
across them. Such a trained model is able to generate forecasts with
little or no history (e.g.~the sales of a not-yet-released item). In our
framework, static features are first embedded into a lower dimensional
representation (a fully-connected layer not shown in Figure 1), and then
replicated as inputs across time.

\hypertarget{application}{\section{Application}\label{application}}

\begin{figure}
\centering
\includegraphics[width=0.46000\textwidth]{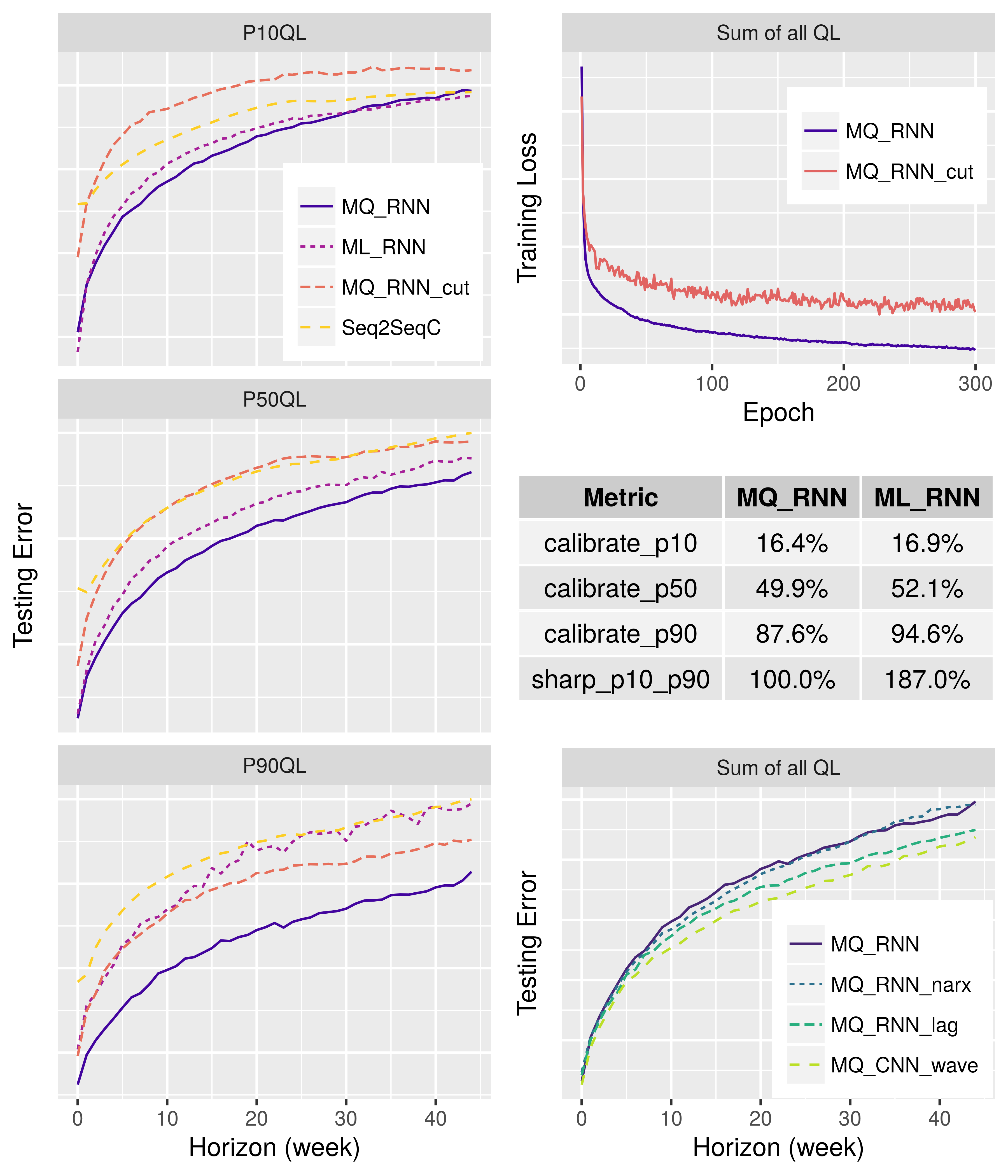}
\caption{Experiment results. \textbf{Left}: Quantile Loss for P10, P50
and P90 forecasts as a function of horizon length. The y-axis is
rescaled and not comparable between panels. \textbf{Upper right}:
training loss versus number of epochs. \textbf{Middle right}:
calibration \(\mathbb{E}(\text{I}(y_{t+k} \leq \hat{y}_{t+k}^{(q)}))\)
and sharpness
\(\mathbb{E}|\hat{y}_{t+k}^{(0.9)} - \hat{y}_{t+k}^{(0.1)}|\) across all
\(t\) and \(k\). The sharpness number is scaled by dividing that of
\texttt{MQ-RNN}. For instance, a perfect calibration for a P90 forecast
is 90\%. If forecast is calibrated, a smaller value of sharpness
(average prediction interval width) is preferrable. \textbf{Lower
right}: sum of test error across all quantiles by different choices of
encoder within the MQ-RNN framework. The y-axis is rescaled. See text
for discussion.}
\end{figure}

Our framework can efficiently forecast millions of time series at
industrial scale and pace. We first apply MQ-RNN to the demand
forecasting problem at Amazon, and design a small-scale experiment to
show how our novelties, i.e.~quantile loss vs likelihood, forking- vs
cutting-sequences and multi-horizon vs recursive, can individually boost
model performance. Improvements by using alternative encoders
(e.g.~MQ-CNN) is also discussed. Next, we apply our forecasting
framework to the Global Energy Forecasting Competition 2014 (GEFCom2014,
\href{http://www.drhongtao.com/gefcom/2014}{Hong et al, 2016}) to
demonstrate that the model is flexible, easy to use and powerful: our
result would have won the 1st place in this competition, without much
tuning.

\subsection{Amazon Demand Forecasting}\label{amazon-demand-forecasting}

\begin{figure}
\centering
\includegraphics[width=0.48000\textwidth]{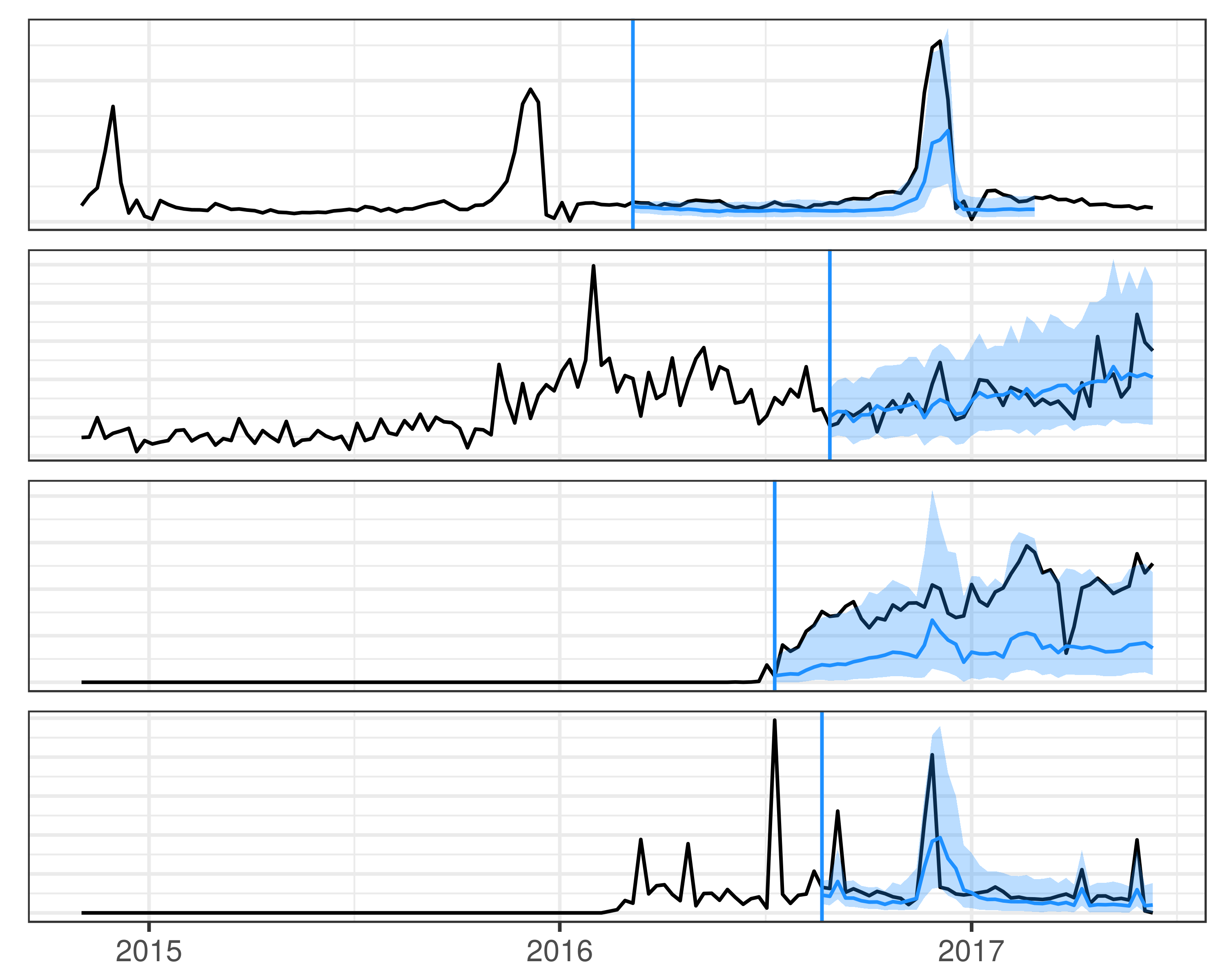}
\caption{MQ-RNN Forecasts for four example products. Dark black line is
the ground truth demand, the vertical line is the forecast creation
time; P10 and P90 forecasts are the lower and upper boundary of the
forecast band, and P50 (median) is the light blue line within the band.
The first two examples are randomly chosen with respectively
long-horizon seasonality and trend; the other two are selected to
illustrate how the model handles new product cold-start situation as
well as promotional spikes.}
\end{figure}

We first describe the dataset we use. Weekly demand series of around
60,000 sampled products from different categories within the US
marketplace are gathered from year 2012 to 2017. Data before 2016 is
used to train the models, and we create multi-horizon forecasts at each
of the 52 weeks in 2016. Forecast horizons range from 1 to 52 weeks.
Available covariates include a range of suitably chosen and standard
demand drivers in three categories: history only, e.g.~past demand;
history and future, e.g.~promotions; and static, e.g.~product catalog
fields. Several models are compared. \texttt{MQ\_RNN} is the proposed
model as in Figure 2b, and other benchmarks are its \emph{minimal}
variants, meaning we modify or knock out a single functionality to mimic
state-of-the-art approaches, while keeping all other
settings/hyper-parameters controlled with best effort. \texttt{ML\_RNN}
changes QL to a shifted Log-Gaussian likelihood:
\(\log{(y+1)} \sim \mathcal{N}(\mu,\sigma^2)\) and predicts
\((\hat{\mu},\hat{\sigma})\); \texttt{MQ\_RNN\_cut} doesn't use
forking-sequences but cuts each series by a FCT; the cut is random
between samples \emph{and} epochs to better use all the information in
the data; \texttt{Seq2SeqC} combines the state-of-the-art Seq2Seq
structure with the predicted Log-Gaussian parameters and recursively
using the one-step-ahead estimated means as inputs for subsequent
forecasts, trained by teacher forcing and cutting-sequences; notably
Seq2SeqC is a more general and efficient benchmark than DeepAR, in that
it is not restricted to an identical LSTM encoder/decoder, and doesn't
need repeated sequential samplings, under the context of estimating the
marginal multi-horizon distributions. In addition, the encoder in
\texttt{MQ\_RNN} can be replaced as described in
\protect\hyperlink{encoder-extensions}{Section 3.4}, resulting in
\texttt{MQ\_RNN\_narx} with the last 52 states skip-connected,
\texttt{MQ\_RNN\_lag} with the last 52 demand values as input, and
\texttt{MQ\_CNN\_wave} with layers of dilated convolutions as the
encoder, respectively. Quantiles are estimated for
\(q\in \{0.1,0.5,0.9\}\) (P10, P50 and P90 forecasts), either directly
or inferred from the Log-Gaussian.

Experiment results are summarized in Figure 3. With all the proposed
structural improvements, \texttt{MQ\_RNN} has consistently the best
accuracy across all horizons. The training loss curve of
\texttt{MQ\_RNN\_cut} is more volatile and flattens out early.
Series-level diagnostics also indicate similar high-level behaviors
between \texttt{MQ\_RNN} and \texttt{MQ\_RNN\_cut}, but the latter has
worse performance. In terms of calibration \texttt{ML\_RNN} is slightly
overbiased, and its 80\% prediction interval is on average almost twice
as wide as \texttt{MQ\_RNN}. We hypothesize that this is because of the
model mis-specification (e.g.~tail behavior) when assuming a
Log-Gaussian on this dataset, and usually further modeling is needed.
The nonparametric quantile regression is robust to this, and both
quantile-based models stand out for P90QL, which focuses on the tail of
the distribution. Contrary to what we expected, \texttt{Seq2SeqC} in
fact has no disadvantage at long-horizon, but its forecast curves are
usually flat. We suspect the Recursive strategy is inducing too much
dependency on the lag mean estimate. By extending the sequential encoder
beyond vanilla LSTM, further accuracy gain is achieved within the
proposed MQ-RNN framework. \texttt{MQ\_RNN\_lag} is the best RNN-type
model, while \texttt{MQ\_CNN\_wave} has the highest accuracy overall,
both with otherwise similar forecast behavior (not shown). Note
\texttt{MQ\_CNN\_wave} is the only model here that is not a minimal
variant of \texttt{MQ\_RNN}, so the comparison could be confounded by
the choice of hyper-parameters. Finally, some anecdotal \texttt{MQ-RNN}
examples are selected and presented in Figure 4, to give readers a
qualitative impression of how the network deals with different use
cases.

\subsection{GEFCom2014 Electricity
Forecasting}\label{gefcom2014-electricity-forecasting}

We also applied MQ-RNN to two external forecasting problems using
datasets published in GEFCom2014 forecasting competition. This
competition had four problems, electricity load forecasting, electricity
price forecasting, and two problems related to wind and solar power
generation. We chose the first two electricity forecasting problems
because 1) they are probabilistic, 2) they are multi-horizon problems,
and 3) they also contain some information about the future horizons. In
this sense the structure of the problems matches quite well the demand
forecasting task. The difference is that the quantity to forecast is a
single series of hour-grain price or load from several years and thus
there is no static series-related information.

Both problems are set with 12 different forecast creation dates. The
competing metric for both is a sum of quantile losses over 99
percentiles of the predicted distributions, and the average loss over
the 12 forecast dates is the final evaluation criterion. In both
problems we trained MQ-RNN to predict quantiles
\(\{0.01, 0.25, 0.5, 0.75, 0.99\}\). Linear interpolation is used to
produce the full set of 99 quantiles.

The electricity price forecasting problem was to forecast hourly price
distributions for a 24-hour horizon (\(24\times 99\) quantile forecasts)
of a particular zone. Information provided about the future consists of
zonal and total load forecasts for the horizon, which were also
available for the past. To this we added calendar-based features about
the day of year and hour in a day, as well as weekday and US holiday
indicators. We would have achieved the 1st place in the competition by
our average quantile loss of 2.63, as opposed to 2.72, 2.73, and 2.82 of
the winner, the 2nd, and 3rd place holders.

\begin{figure}
\centering
\includegraphics[width=0.48000\textwidth]{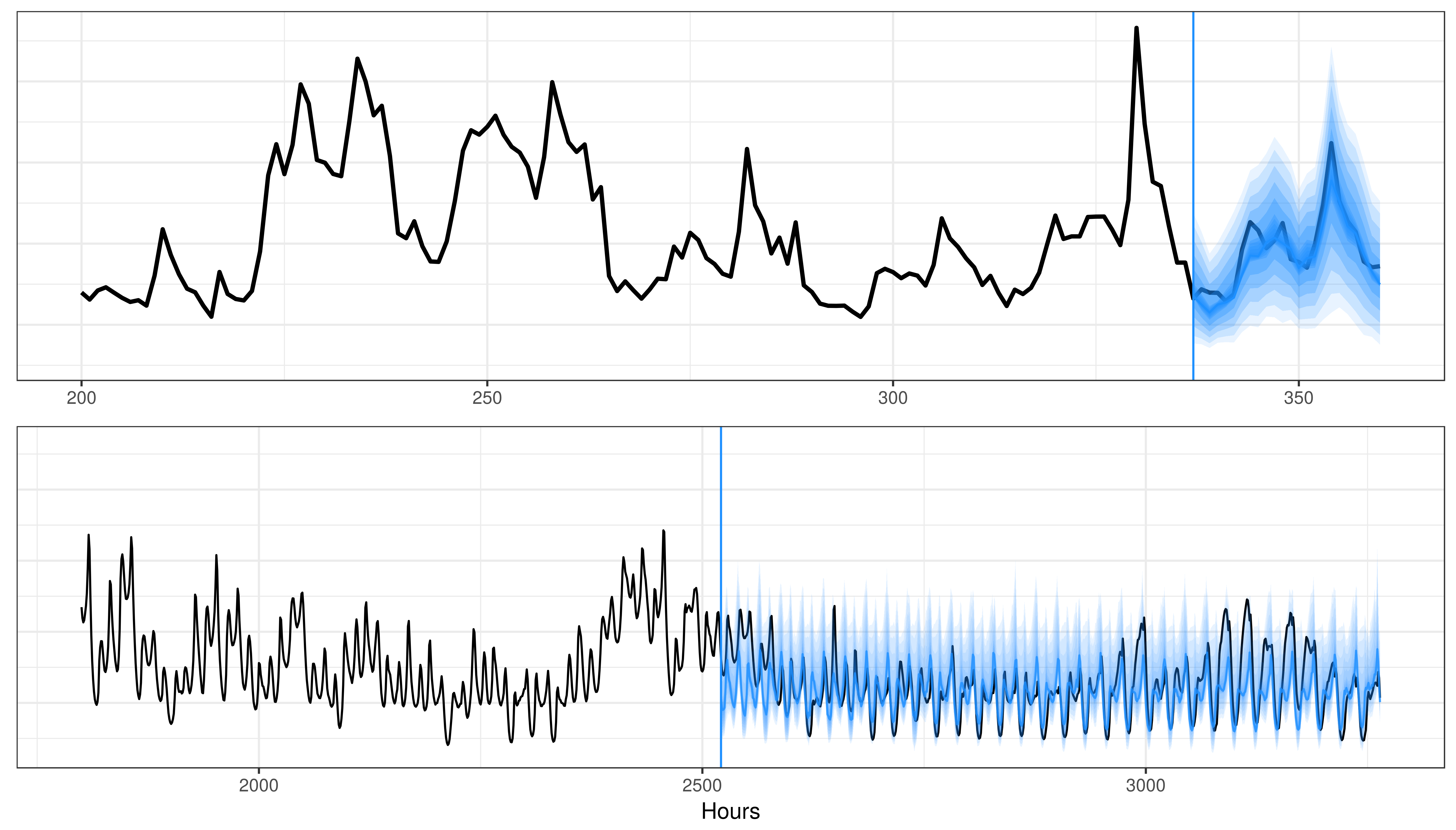}
\caption{MQ-RNN Forecasts for GEFCom 2014 dataset. \textbf{Upper}:
electricity price, task 12, to forecast 24 hours ahead. \textbf{Lower}:
electricity load, task 4, to forecast 744 hours ahead. Dark black line
is ground truth, the vertical line is forecast creation time; Forecast
bands are plotted for every other 5 percentiles, i.e. {[}P1,P99{]},
{[}P5,P95{]}, {[}P10,P90{]}, etc. P50 (median) is the solid blue line
within the band. For clarity, past series is not shown in full length.}
\end{figure}

The electricity load forecasting problem calls for forecasts of hourly
load distributions of a certain US utility for a month into the future
(\(744\times 99\) quantile forecasts). In this case the future
information is solely calendar-based. Weather was available for the past
as temperature measurements of 25 weather stations. In order to capture
longer time dynamics without too long RNNs, we chose to run the encoder
at a daily grain, keeping the forecasting decoder grain as hours. In
this problem we would have won as well, achieving average quantile loss
of 7.43. The top three competitors had quantile losses of 7.45, 7.51,
and 7.83.

The networks are not intensively tuned, and the final setting is based
on intuitive first few tries. The major parameter choices are the
duration of the time-steps that the RNN is modeling (number of recurrent
layers) and the number of RNN states. These parameters determine the
dynamics of the history captured by the RNN hidden state. For the price
prediction task we chose 168 hours as the duration, and for the load
prediction 56 days, both with a state dimension of 30. For training,
mini-batches are random slices of the multi-year past data such that the
durations match our choice of RNN length, and we train with
forking-sequences for each slice as a sample. For each of the 12
forecast creation dates we use data prior the date for training, and
then retrain from scratch for each subsequent forecast creation date. We
also used the lag-series trick as \texttt{MQ\_RNN\_lag} in the previous
subsection: the RNN input at time \(t\) is not only the time-series
value at \(t\) but a vector of lagged values of 168 past hours for
price, and 7 days for load. Figure 5 shows example forecasts from each
of the problem.

\hypertarget{conclusion}{\section{Conclusion}\label{conclusion}}

We presented a general framework for probabilistic time series
regression, and demonstrated how the novel components can each
contribute to the final performance over state-of-arts. Our findings can
help in the design of both practical large-scale forecasting
applications and encoder-decoder style deep learning architecture. In
this work, we have not discussed some extensions, including explicit
multivariate forecasting and modeling the joint distribution of
horizons. These will be addressed in future texts.

\subsubsection*{Acknowledgment}\label{acknowledgment}
\addcontentsline{toc}{subsubsection}{Acknowledgment}

We would like to thank Ping Xu and Charles Elkan for their helpful
comments and support.

\section*{Reference}\label{reference}
\addcontentsline{toc}{section}{Reference}

Bengio, S., Vinyals, O., Jaitly, N., \& Shazeer, N. (2015). Scheduled
sampling for sequence prediction with recurrent neural networks. In
Advances in Neural Information Processing Systems (pp.~1171-1179).

Bianchi, F. M., Maiorino, E., Kampffmeyer, M. C., Rizzi, A., \& Jenssen,
R. (2017). An overview and comparative analysis of recurrent neural
networks for short term load forecasting. arXiv preprint
arXiv:1705.04378.

Bishop, C. M. (1994). Mixture density networks.

Borovykh, A., Bohte, S., \& Oosterlee, C. W. (2017). Conditional time
series forecasting with convolutional neural networks. arXiv preprint
arXiv:1703.04691.

Box, G. E., Jenkins, G. M., Reinsel, G. C., \& Ljung, G. M. (2015). Time
series analysis: forecasting and control. John Wiley \& Sons.

Chevillon, G. (2007). Direct multi‐step estimation and forecasting.
Journal of Economic Surveys, 21(4), 746-785.

Cho, K., Van Merriënboer, B., Gulcehre, C., Bahdanau, D., Bougares, F.,
Schwenk, H., \& Bengio, Y. (2014). Learning phrase representations using
RNN encoder-decoder for statistical machine translation. arXiv preprint
arXiv:1406.1078.

Cinar, Y. G., Mirisaee, H., Goswami, P., Gaussier, E., Aït-Bachir, A.,
\& Strijov, V. (2017, November). Position-based content attention for
time series forecasting with sequence-to-sequence rnns. In International
Conference on Neural Information Processing (pp.~533-544). Springer,
Cham.

Connor, J., Atlas, L. E., \& Martin, D. R. (1992). Recurrent networks
and NARMA modeling. In Advances in Neural Information Processing Systems
(pp.~301-308).

DiPietro, R., Navab, N., \& Hager, G. D. (2017). Revisiting NARX
Recurrent Neural Networks for Long-Term Dependencies. arXiv preprint
arXiv:1702.07805.

Elman, J. L. (1990). Finding structure in time. Cognitive science,
14(2), 179-211.

Flunkert, V., Salinas, D., \& Gasthaus, J. (2017). DeepAR: Probabilistic
forecasting with autoregressive recurrent networks. arXiv preprint
arXiv:1704.04110.

Gers, F. A., Schmidhuber, J., \& Cummins, F. (1999). Learning to forget:
Continual prediction with LSTM.

Graves, A. (2013). Generating sequences with recurrent neural networks.
arXiv preprint arXiv:1308.0850.

Hong, T., Pinson, P., Fan, S., Zareipour, H., Troccoli, A., \& Hyndman,
R. J. (2016). Probabilistic energy forecasting: Global energy
forecasting competition 2014 and beyond. International Journal of
Forecasting, vol.32, no.3 (pp 896-913).

Koenker, R., \& Bassett Jr, G. (1978). Regression quantiles.
Econometrica: journal of the Econometric Society, 33-50.

Lamb, A. M., GOYAL, A. G. A. P., Zhang, Y., Zhang, S., Courville, A. C.,
\& Bengio, Y. (2016). Professor forcing: A new algorithm for training
recurrent networks. In Advances In Neural Information Processing Systems
(pp.~4601-4609).

Längkvist, M., Karlsson, L., \& Loutfi, A. (2014). A review of
unsupervised feature learning and deep learning for time-series
modeling. Pattern Recognition Letters, 42, 11-24.

Lipton, Z. C., Kale, D. C., Elkan, C., \& Wetzel, R. (2015). Learning to
diagnose with LSTM recurrent neural networks. arXiv preprint
arXiv:1511.03677.

Ng, N., Gabriel, R. A., McAuley, J., Elkan, C., \& Lipton, Z. C. (2017).
Predicting Surgery Duration with Neural Heteroscedastic Regression.
arXiv preprint arXiv:1702.05386.

Taieb, S. B., \& Atiya, A. F. (2016). A bias and variance analysis for
multistep-ahead time series forecasting. IEEE transactions on neural
networks and learning systems, 27(1), 62-76.

Taylor, J. W. (2000). A quantile regression neural network approach to
estimating the conditional density of multiperiod returns. Journal of
Forecasting, 19(4), 299-311.

Van Den Oord, A., Dieleman, S., Zen, H., Simonyan, K., Vinyals, O.,
Graves, A., \ldots{} \& Kavukcuoglu, K. (2016). Wavenet: A generative
model for raw audio. arXiv preprint arXiv:1609.03499.

Xu, Q., Liu, X., Jiang, C., \& Yu, K. (2016). Quantile autoregression
neural network model with applications to evaluating value at risk.
Applied Soft Computing, 49, 1-12.

\end{document}